\pdfoutput=1

\documentclass[letterpaper]{article}
\usepackage{proceed2e}
\usepackage[margin=1in]{geometry}

\usepackage{times}

\usepackage{graphicx}
\usepackage{subfig}
\usepackage{algorithm2e}
\RestyleAlgo{boxruled}
\usepackage{amsthm}
\usepackage{amsmath}
\usepackage{amssymb}
\usepackage{xcolor}
\usepackage{natbib}
\usepackage{float}
\usepackage[bottom]{footmisc}

\newtheorem{theorem}{Theorem}
\newcommand{\argmax}{\text{arg}\max}

\usepackage{lipsum}
\usepackage{fancyhdr}
\title{A Practical Method for Solving Contextual Bandit \\ Problems Using Decision Trees}

\author{ {\bf Adam N. Elmachtoub}\thanks{\, Authors are listed alphabetically.}  \\
IEOR Department          \\
Columbia University\\
\And
{\bf Ryan McNellis} \\
IEOR Department \\
Columbia University\\
\And
{\bf Sechan Oh}   \\
Moloco \\
Palo Alto, CA 94301 \\
\And
{\bf Marek Petrik}   \\
Dept. of Computer Science \\
University of New Hampshire    \\
}

\newcommand{\Beta}{\text{Beta}}

\begin{document}

\thispagestyle{fancy}

\maketitle

\begin{abstract}
Many efficient algorithms with strong theoretical guarantees have been proposed for the contextual multi-armed bandit problem. However, applying these algorithms in practice can be difficult because they require domain expertise to build appropriate features and to tune their parameters. We propose a new method for the contextual bandit problem that is simple, practical, and can be applied with little or no domain expertise. Our algorithm relies on decision trees to model the context-reward relationship. Decision trees are non-parametric, interpretable, and work well without hand-crafted features. To guide the exploration-exploitation trade-off, we use a bootstrapping approach which abstracts Thompson sampling to non-Bayesian settings. We also discuss several computational heuristics and demonstrate the performance of our method on several datasets. 
\end{abstract}

\section{INTRODUCTION}

Personalized recommendation systems play a fundamental role in an ever-increasing array of settings. For example, many mobile and web services earn revenue primarily from targeted advertising \citep{yuan2003recommendation, dhar2011challenges}. The effectiveness of this strategy is predicated upon intelligently assigning the right ads to users based on contextual information. Other examples include assigning treatments to medical patients \citep{kim2011battle} and recommending web-based content such as news articles to subscribers \citep{li2010contextual}.

In this paper, we tackle the problem where little or no prior data is available, and an algorithm is required to ``learn" a good recommendation system in real time as users arrive and data is collected. This problem, known as the \textit{contextual bandit problem} (or contextual multi-armed bandit problem), relies on an algorithm to navigate the \textit{exploration-exploitation} trade-off when choosing recommendations. Specifically, it must simultaneously exploit knowledge from data accrued thus far to make high-impact recommendations, while also exploring recommendations which have a high degree of uncertainty in order to make better decisions in the future. 

The contextual bandit problem we consider can be formalized as follows \citep{langford2008epoch}. At each time point, a user arrives and we receive a vector of information, henceforth referred to as the \textit{context}. We must then choose one of $K$ distinct \textit{actions} for this user. Finally, a random outcome or \textit{reward} is observed, which is dependent on the user and action chosen by the algorithm. Note that the probability distribution governing rewards for each action is unknown and depends on the observed context. The objective is to maximize the total rewards accrued over time. In this paper, we focus on settings in which rewards are binary, observing a \textit{success} or a \textit{failure} in each time period. However, our algorithm does not rely on this assumption and may also be used in settings with continuous reward distributions.


A significant number of algorithms have been proposed for the contextual bandit problem, often with strong theoretical guarantees \citep{auer2002using, filippi2010parametric,chu2011contextual,hsu2014taming}. However, we believe that many of these algorithms cannot be straightforwardly and effectively applied in practice for personalized recommendation systems, as they tend to exhibit at least one of the following drawbacks.

\begin{enumerate}
	\item \textit{Parametric modeling assumptions}. The vast majority of bandit algorithms assume a parametric relationship between contexts and rewards \citep{li2010contextual,filippi2010parametric,abbasi2011improved,agrawal2013thompson}. To use such algorithms in practice, one must first do some feature engineering and transform the data to satisfy the parametric modeling framework. However, in settings where very little prior data is available, it is often unclear how to do so in the right way.
	
	\item \textit{Unspecified constants in the algorithm}. Many bandit algorithms contain unspecified parameters which are meant to be tuned to control the level of exploration \citep{auer2002using,li2010contextual,filippi2010parametric,allesiardo2014neural}. Choosing the wrong parameter values can negatively impact performance \citep{russo2014learning}, yet choosing the right values is difficult since little prior data is available. 
	
	\item \textit{Ill-suited learners for classification problems}. It is commonly assumed in the bandit literature that the relationship between contexts and rewards is governed by a linear model \citep{li2010contextual,abbasi2011improved,agrawal2013thompson}. Although such models work well in regression problems, they often face a number of issues when estimating probabilities from binary response data \citep{scott1997regression}.
	
\end{enumerate}

In the hopes of addressing all of these issues, we propose a new algorithm for the contextual bandit problem which we believe can be more effectively applied in practice. Our approach uses decision tree learners to model the context-reward distribution for each action. Decision trees have a number of nice properties which make them effective, requiring no data modification or user input before being fit \citep{friedman2001elements}. To navigate the exploration-exploitation tradeoff, we use a parameter-free bootstrapping technique that emulates the core principle behind Thompson sampling. We also provide a computational heuristic to improve the speed of our algorithm. Our simple method works surprisingly well on a wide array of simulated and real-world datasets compared to well-established algorithms for the contextual bandit problem.

\section{LITERATURE REVIEW}

Most contextual bandit algorithms in the existing literature can be categorized along two dimensions: (i) the base learner and (ii) the exploration algorithm. Our method uses decision tree learners in conjunction with bootstrapping to handle the exploration-exploitation trade-off. To the best of our knowledge, the only bandit algorithm which applies such learners is BanditForest \citep{feraud2016random}, using random forests as the base learner with decision trees as a special case. One limitation of the algorithm is that it depends on four problem-specific parameters requiring domain expertise to set: two parameters directly influence the level of exploration, one controls the depth of the trees, and one determines the number of trees in the forest. By contrast, our algorithm requires no tunable parameters in its exploration and chooses the depth internally when building the tree. Further, BanditForest must sample actions uniformly-at-random until all trees are completely learned with respect to a particular context. As our numerical experiments show, this leads to  excessive selection of low-reward actions, causing the algorithm's empirical performance to suffer. NeuralBandit \citep{allesiardo2014neural} is an algorithm which uses neural networks as the base learner. Using a probability specified by the user, it randomly decides whether to explore or exploit in each step. However, choosing the right probability is rather difficult in the absence of data.

Rather than using non-parametric learners such as decision trees and neural nets, the vast majority of bandit algorithms assume a parametric relationship between contexts and rewards. Commonly, such learners assume a monotonic relationship in the form of a linear model \citep{li2010contextual,abbasi2011improved,agrawal2013thompson} or a Generalized Linear Model \citep{filippi2010parametric, li2011unbiased}. However, as with all methods that assume a parametric structure in the context-reward distribution, manual transformation of the data is often required in order to satisfy the modeling framework. In settings where little prior data is available, it is often quite difficult to ``guess" what the right transformation might be. Further, using linear models in particular can be problematic when faced with binary response data, as such methods can yield poor probability estimates in this setting \citep{scott1997regression}. 

Our method uses bootstrapping in a way which ``approximates" the behavior of Thompson sampling, an exploration algorithm which has recently been applied in the contextual bandit literature \citep{agrawal2013thompson, russo2014learning}. Detailed in Section \ref{sec:ThompsonSampling}, Thompson sampling is a Bayesian framework requiring a parametric response model, and thus cannot be straightforwardly applied when using decision tree learners. The connection between bootstrapping and Thompson sampling has been previously explored in \citet{eckles2014thompson}, \citet{osband2015bootstrapped}, and \citet{tang2015personalized}, although these papers either focus on the context-free case or only consider using parametric learners with their bootstrapping framework. \citet{baransi2014sub} propose a sub-sampling procedure which is related to Thompson sampling, although the authors restrict their attention to the context-free case.

Another popular exploration algorithm in the contextual bandit literature is Upper Confidence Bounding (UCB) \citep{auer2002using,li2010contextual,filippi2010parametric}. These methods compute confidence intervals around expected reward estimates and choose the action with the highest upper confidence bound. UCB-type algorithms often rely on a tunable parameter controlling the width of the confidence interval. Choosing the right parameter value can have a huge impact on performance, but with little problem-specific data available this can be quite challenging \citep{russo2014learning}. Another general exploration algorithm which heavily relies on user input is Epoch-Greedy \citep{langford2008epoch}, which depends on an unspecified (non-increasing) function to modulate between exploration and exploitation. 

Finally, a separate class of contextual bandit algorithms are those which select the best policy from an exogenous finite set, such as \citet{dudik2011efficient} and \citet{hsu2014taming}. The performance of such algorithms depends on the existence of a policy in the set which performs well empirically. In the absence of prior data, the size of the required set may be prohibitively large, resulting in poor empirical and computational performance. Furthermore, in a similar manner as UCB, these algorithms require a tunable parameter that influences the level of exploration.

\section{PROBLEM FORMULATION}

In every time step $t=1,\ldots, T$, a user arrives with an $M$-dimensional context vector $x_t \in \mathbb{R}^M$. Using $x_t$ as input, the algorithm chooses one of $K$ possible actions for the user at time $t$. We let $a_t \in \{1,...,K\}$ denote the action chosen by the algorithm. We then observe a random, binary reward $r_{t,a_t} \in \{0,1\}$ associated with the chosen action $a_t$ and context $x_t$. Our objective is to maximize the cumulative reward over the time horizon.

Let $p(a,x)$ denote the (unknown) probability of observing a positive reward, given that we have seen context vector $x$ and offered action $a$. We define the \textit{regret} incurred at time $t$ as 
\[ \max_a E[r_{t,a}] - E[r_{t,a_t}] = \max_a p(a, x_t) - p(a_t, x_t)\,. \]

Intuitively, the regret measures the expected difference in reward between a candidate algorithm and the best possible assignment of actions. An equivalent objective to maximizing cumulative rewards is to minimize the $T$-period \textit{cumulative regret}, $R(T)$, which is defined as

\begin{equation*}
R(T) = \sum_{t=1}^T \max_a p(a, x_t) - p(a_t, x_t)\,.
\end{equation*}

In our empirical studies, we use cumulative regret as the performance metric for assessing our algorithm.

\section{PRELIMINARIES} 

\subsection{DECISION TREES} \label{sec:DecisionTrees}

\begin{figure}[h]
	\center{\includegraphics[width=0.45\textwidth]{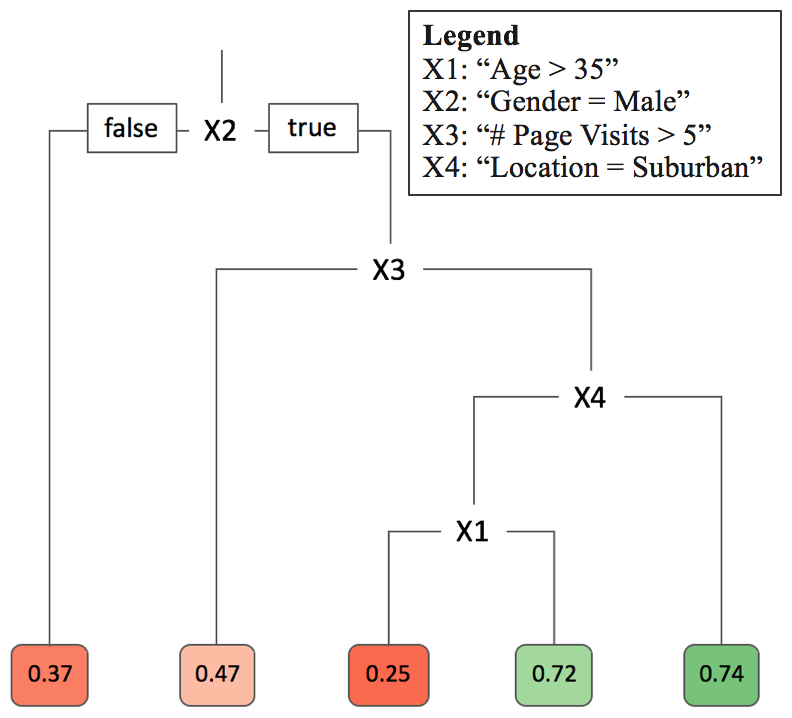}}
	\caption{A decision tree modeling the distribution of rewards for a golf club advertisement. Terminal nodes display the success probability corresponding to each group of users.}
	\label{fig:DecisionTree}
\end{figure}

Our method uses decision tree learners in modeling the relationship between contexts and success probabilities for each action. Decision trees have a number of desirable properties: they handle both continuous and binary response data efficiently, are robust to outliers, and do not rely on any parametric assumptions about the response distribution. Thus, little user configuration is required in preparing the data before fitting a decision tree model. They also have the benefit of being highly interpretable, yielding an elegant visual representation of the relationship between contexts and rewards \citep{friedman2001elements}. 
Figure \ref{fig:DecisionTree} provides a diagram of a decision tree model for a particular sports advertisement. Observe that the tree partitions the context space into different regions, referred to as \textit{terminal nodes} or \textit{leaves}, and we assume that each of the users belonging to a certain leaf has the same success probability. 

There are many efficient algorithms proposed in the literature for estimating decision tree models from training data. In our numerical experiments, we used the CART algorithm with pruning as described by \citet{breiman1984classification}. Note that this method does not rely on any tunable parameters, as the depth of the tree is selected internally using cross-validation.

\subsection{THOMPSON SAMPLING FOR CONTEXTUAL BANDITS} \label{sec:ThompsonSampling}

Our algorithm was designed to mimic the behavior of Thompson sampling, a general algorithm for handling the exploration-exploitation trade-off in bandit problems. In addition to having strong performance guarantees relative to UCB \citep{russo2014learning}, Thompson sampling does not contain unspecified constants which need to be tuned for proper exploration. Thus, there is sufficient motivation for using this method in our setting.

In order to provide intuition for our algorithm, we describe how Thompson sampling works in the contextual bandit case. To begin, assume that each action $a$ has a set of unknown parameters $\theta_a$ governing its reward distribution, $P(r_a | \theta_a, x_t)$. For example, when using linear models with Thompson sampling, $E[r_a | \theta_a, x_t] = x_t' \theta_a$. We initially model our uncertainty about $\theta_a$ using a pre-specified prior distribution, $P(\theta_a)$. As new rewards are observed for action $a$, we update our model accordingly to its so-called posterior distribution, $P(\theta_a | D_{t,a})$. Here, $D_{t,a}$ represents the set of context/reward pairs corresponding to times up to $t$ that action $a$ was offered, i.e., $D_{t,a} = \{(x_s,r_{s,a}) : s \leq t-1, a_s = a\}$.

Thompson sampling behaves as follows. During each time step, every action's parameters $\theta_a$ are first randomly sampled according to posterior distribution $P(\theta_a | D_{t,a})$. Then, Thompson sampling chooses the action which maximizes expected reward with respect to the sampled parameters. In practice, this can be implemented according to the pseudocode given in Algorithm \ref{alg:ThompsonSampling}.

\begin{algorithm}

	\caption{ThompsonSampling()}
	\For{$t=1,...,T$}{
	Observe context vector $x_t$\\
	\For{$a=1,...,K$}{
			Sample $\tilde{\theta}_{t,a}$ from $P(\theta_a | D_{t,a})$
	}
	Choose action $a_t = \displaystyle{\argmax_{a} \  E[r_a|\tilde{\theta}_{t,a},x_t]}$\\
	Update $D_{t,a_t}$ and $P(\theta_{a_t} | D_{t,a_t})$ with $(x_t, r_{t,a_t})$ \\
	
	} \label{alg:ThompsonSampling}
\end{algorithm} 

\section{THE BOOTSTRAPPING ALGORITHM}

\vspace{5 mm}

\subsection{BOOTSTRAPPING TO CREATE A ``THOMPSON SAMPLE"}

As decision trees are inherently non-parametric, it is not straightforward to mathematically define priors and posteriors with respect to these learners, making Thompson sampling difficult to implement. However, one can use bootstrapping to simulate the behavior of sampling from a posterior distribution, the intuition for which is discussed below.

Recall that data $D_{t,a}$ is the set of observations for action $a$ at time $t$. If we had access to many i.i.d. datasets of size $|D_{t,a}|$ for action $a$, we could fit a decision tree to each one and create a \textit{distribution} of trees. A Thompson sample could then be generated by sampling a random tree from this collection. Although assuming access to these datasets is clearly impractical in our setting, we can use bootstrapping to approximate this behavior. We first create a large number of bootstrapped datasets, each one formed by sampling $|D_{t,a}|$ context-reward pairs from $D_{t,a}$ with replacement. Then, we fit a decision tree to each of these bootstrapped datasets. A ``Thompson sample'' is then a randomly selected tree from this collection. Of course, an equivalent and computationally more efficient procedure is to simply fit a decision tree to a \textit{single} bootstrapped dataset. This intuition serves as the basis for our algorithm.

Following \citet{tang2015personalized}, let $\tilde{D}_{t,a}$ denote the dataset obtained by bootstrapping $D_{t,a}$, i.e. sampling $|D_{t,a}|$ observations from $D_{t,a}$ with replacement. Denote the decision tree fit on $\tilde{D}_{t,a}$ by $\tilde{\theta}_{t,a}$. Finally, let $\hat{p}(\tilde{\theta},x_t)$ denote the estimated probability of success from using decision tree $\tilde{\theta}$ on context $x_t$.

At each time point, the bootstrapping algorithm simply selects the action which maximizes $\hat{p}(\tilde{\theta}_{t,a},x_t)$. See Algorithm \ref{alg:TreeBootstrap} for the full pseudocode. Although our method is given with respect to decision tree models, note that this bootstrapping framework can be used with any base learner, such as logistic regression, random forests, and neural networks. 

\begin{algorithm}
	\caption{TreeBootstrap()}
	\For{$t=1,...,T$}{
		Observe context vector $x_t$\\
		\For{$a=1,...,K$}{
			Sample bootstrapped dataset $\tilde{D}_{t,a}$ from $D_{t,a}$\\
			Fit decision tree $\tilde{\theta}_{t,a}$ to $\tilde{D}_{t,a}$
		}
		Choose action $a_t = \displaystyle{\argmax_{a} \ \hat{p}(\tilde{\theta}_{t,a},x_t)}$\\
		Update $D_{t,a_t}$ with $(x_t, r_{t,a_t})$ \\
	}  
	\label{alg:TreeBootstrap}
	
\end{algorithm} 

Observe that TreeBootstrap may eliminate an action after a single observation if its first realized reward is a failure, as the tree constructed from the resampled dataset in subsequent iterations will always estimate a success probability of zero. There are multiple solutions to address this issue. First, one can force the algorithm to continue offering each action $a$ until a success is observed (or an action $a$ is eliminated after a certain threshold number of failures). This is the approach used in our numerical experiments. Second, one can add fabricated prior data of one success and one failure for each action, where the associated context for the prior is that of the first data point observed. The prior data prohibit the early termination of arms, and their impact on the prediction accuracy becomes negligible as the number of observations increases. 

\subsection{MEASURING THE SIMILARITY BETWEEN BOOTSTRAPPING AND THOMPSON SAMPLING}
Here we provide a simple result that quantifies how close the Thompson sampling and bootstrapping algorithms are in terms of the actions chosen in each time step. Measuring the closeness of the two algorithms in the contextual bandit framework is quite challenging, and thus we focus on the standard (context-free) multi-armed bandit problem in this subsection. Suppose that the reward from choosing each action, i.e., $r_{t,a}$, follows a Bernoulli distribution with an unknown success probability. At a given time $t$, let $n_a$ denote the total number of times that action $a$ has been chosen, and let $p_a$ denote the proportion of successes observed for action $a$.

In standard multi-armed bandits with Bernoulli rewards, Thompson sampling first draws a random sample of the true (unknown) success probability for each action $a$ according to a Beta distribution with parameters $\alpha_a = n_a p_a$ and $\beta_a = n_a (1-p_a)$. It then chooses the action with the highest sampled success probability. Conversely, the bootstrapping algorithm samples $n_a$ observations with replacement from action $a$'s observed rewards, and the generated success probability is then the proportion of successes observed in the bootstrapped dataset. Note that this procedure is equivalent to generating a binomial random variable with number of trials $n_a$ and success rate $p_a$, divided by $n_a$.

In Theorem 1 below, we bound the difference in the probability of choosing action $a$ when using bootstrapping versus Thompson sampling. For simplicity, we assume that we have observed at least one success and one failure for each action, i.e. $n_a \geq 2$ and $p_a \in (0,1)$ for all $a$.

\begin{theorem}
Let $a_t^{TS}$ be the action chosen by the Thompson sampling algorithm, and let $a_t^{B}$ be the action chosen by the bootstrapping algorithm given data $(n_a,p_a)$ for each action $a \in \{1,2,\ldots,K\}$. Then, 
\[
| P(a_t^{TS} = a) - P(a_t^{B} = a) | \leq C_a(p_1,...,p_K) \sum_{a=1}^K \frac{1}{\sqrt{n_a}}
\]
holds for every $a \in \{1,2,\ldots,K\}$, for some function $C_a(p_1,...,p_K)$ of $p_1,...,p_K$. 
\label{thm:bound}
\end{theorem}
Note that both algorithms will sample each action infinitely often, i.e. $n_a \rightarrow \infty$ as $t \rightarrow \infty$ for all $a$. Hence, as the number of time steps increases, the two exploration algorithms choose actions according to increasingly similar probabilities. Theorem \ref{thm:bound} thus sheds some light onto how quickly the two algorithms converge to the same action selection probabilities. A full proof is provided in the Supplementary Materials section.

\subsection{EFFICIENT HEURISTICS}

Note that TreeBootstrap requires fitting $K$ decision trees from scratch at each time step. Depending on the method used to fit the decision trees as well as the size of $M$, $K$, and $T$, this can be quite computationally intensive. Various online algorithms have been proposed in the literature for training decision trees, referred to as Incremental Decision Trees (IDTs) \citep{crawford1989extensions,utgoff1989incremental,utgoff1997decision}. Nevertheless, Algorithm \ref{alg:TreeBootstrap} does not allow for efficient use of IDTs, as the bootstrapped dataset for an action significantly changes at each time step. 

However, one could modify Algorithm \ref{alg:TreeBootstrap} to instead use an \textit{online} method of bootstrapping. \citet{eckles2014thompson} propose a different bootstrapping framework for bandit problems which is more amenable to online learning algorithms. Under this framework, we begin by initializing $B$ null datasets for every action, and new context-reward pairs are added in real time to each dataset with probability $1/2$. We then simply maintain $K \times B$ IDTs fit on each of the datasets, and a ``Thompson sample" then corresponds to randomly selecting one of an action's $B$ IDTs. Note that there is an inherent trade-off with respect to the number of datasets per action, as larger values of $B$ come with both higher approximation accuracy to the original bootstrapping framework and increased computational cost.

We now propose a heuristic which only requires maintaining one IDT per action, as opposed to $K \times B$ IDTs. Moreover, only one tree update is needed per time period. The key idea is to simply maintain an IDT $\hat{\theta}_{t,a}$ fit on each action's dataset $D_{t,a}$. Then, using the leaves of the trees to partition the context space into regions, we treat each region as a standard multi-arm bandit (MAB) problem. More specifically, let $N_1(\hat{\theta}_{t,a},x_t)$ denote the number of successes in the leaf node of $\hat{\theta}_{t,a}$ corresponding to $x_t$, and analogously define $N_0(\hat{\theta}_{t,a},x_t)$ as the number of failures. Then, we simply feed this data into a standard MAB algorithm, where we assume action $a$ has observed $N_1(\hat{\theta}_{t,a},x_t)$ successes and $N_0(\hat{\theta}_{t,a},x_t)$ failures thus far. Depending on the action we choose, we then update the corresponding IDT of that action and proceed to the next time step. 

Algorithm \ref{alg:TreeHeuristic} provides the pseudocode for this heuristic using the standard Thompson sampling algorithm for multi-arm bandits. Note that this requires a prior number of successes and failures for each context region, $S_0$ and $F_0$. In the absence of any problem-specific information, we recommend using the uniform prior $S_0 = F_0 = 1$. 

\begin{algorithm} 
	\caption{TreeHeuristic()}
	\For{$t=1,...,T$}{
		Observe context vector $x_t$\\
		\For{$a=1,...,K$}{
			Sample $TS_{t,a} \sim \Beta(N_1(\hat{\theta}_{t,a},x_t) + S_0, N_0(\hat{\theta}_{t,a},x_t) + F_0)$
		}
		Choose action $a_t = \displaystyle{\argmax_{a} \  TS_{t,a}}$\\
		Update tree $\hat{\theta}_{t,a_t}$ with $(x_t, r_{t,a_t})$ \\
	}
	\label{alg:TreeHeuristic}
\end{algorithm}

Both TreeBootstrap and TreeHeuristic are algorithms which aim to emulate the behavior of Thompson Sampling. TreeHeuristic is at least $O(K)$ times faster computationally than TreeBootstrap, as it only requires refitting one decision tree per time step -- the tree corresponding to the sampled action. However, TreeHeuristic sacrifices some robustness in attaining these computational gains. In each iteration of TreeBootstrap, a new decision tree is resampled for every action to account for two sources of uncertainty: (a) global uncertainty in the tree structure (i.e. are we splitting on the right variables?) and (b) local uncertainty in each leaf node (i.e., are we predicting the correct probability of success in each leaf?). Conversely, TreeHeuristic keeps the tree structures fixed and only resamples the data in leaf nodes corresponding to the current context -- thus, TreeHeuristic only accounts for uncertainty (b), not (a). Note that if both the tree structures \textit{and} the leaf node probability estimates were kept fixed, this would amount to a pure exploitation policy.


\section{EXPERIMENTAL RESULTS}

We assessed the empirical performance of our algorithm using the following sources of data as input:

\begin{enumerate}
	\item A simulated ``sports ads" dataset with decision trees for each offer governing reward probabilities.
	
	\item Four classification datasets obtained from the UCI Machine Learning Repository \citep{Lichman:2013}: \textit{Adult}, \textit{Statlog (Shuttle)}, \textit{Covertype}, and \textit{US Census Data (1990)}.
\end{enumerate}

We measured the cumulative regret incurred by TreeBootstrap on these datasets, and we compare its performance with that of TreeHeuristic as well as several benchmark algorithms proposed in the bandit literature.

\subsection{BENCHMARK ALGORITHMS}

We tested the following benchmarks in our computational experiments:
\begin{enumerate}
\item\textit{Context-free MAB}. To demonstrate the value of using contexts in  recommendation systems, we include the performance of a context-free multi-arm bandit algorithm as a benchmark. Specifically, we use context-free Thompson sampling in our experiments.
	
\item \textit{BanditForest}. To the best of our knowledge, BanditForest is the only bandit algorithm in the literature which uses decision tree learners. Following the approach used in their numerical studies, we first recoded each continuous variable into five binary variables before calling the algorithm. Note this is a necessary preprocessing step, as the algorithm requires all contexts to be binary. The method contains two tunable parameters which control the level of exploration, $\delta$ and $\epsilon$, which we set to the values tested in their paper: $\delta = 0.05$, and $\epsilon \sim \text{Uniform}(0.4,0.8)$. Additionally, two other tunable parameters can be optimized: the depth of each tree, $D$, and the number of trees in the random forest, $L$. We report the values of these parameters which attained the best cumulative regret with respect to our time horizon: $L = 3$ and $D = 4, 2, 4, 5,$ and $2$ corresponding to the simulated sports-advertising dataset, \textit{Adult}, \textit{Shuttle}, \textit{Covertype}, and \textit{Census}, respectively. Note that the optimal parameter set varies depending on the dataset used. In practice, one cannot know the optimal parameter set in advance without any prior data, and so the performance we report may be optimistic.
	
\item \textit{LinUCB}. Developed by \citet{li2010contextual}, LinUCB is one of the most cited contextual bandit algorithms in the recent literature. The method calls for fitting a ridge regression model on the context-reward data for each action (with regularization parameter $\lambda = 1$). We then choose the action with the highest upper confidence bound with respect to a new context's estimated probability of success. All contexts were scaled to have mean 0 and variance 1 before calling the algorithm, and all categorical variables were binarized. Due to the high-dimensionality of our datasets (particularly after binarization), the predictive linear model requires sufficient regularization to prevent overfitting. In the hopes of a fairer comparison with our algorithm, we instead select the regularization parameter from a grid of candidate values using cross-validation. We report the best cumulative regret achieved when varying the UCB constant $\alpha$ among a grid of values from 0.0001 to 10. Similar to the BanditForest case, the optimal parameter depends on the dataset.
	
\item \textit{LogisticUCB}. As we are testing our algorithms using classification data, there is significant motivation to use a logistic model, as opposed to a linear model, in capturing the context-reward relationship. \citet{filippi2010parametric} describe a bandit algorithm using a generalized linear modeling framework, of which logistic regression is a special case. However, the authors tackle a problem formulation which is slightly different than our own. In their setting, each \textit{action} has an associated, non-random context vector $x_a$, and the expected reward is a function of a single unknown parameter $\theta: E[r_t | x_a] = \mu(x_a^T \theta)$ (here, $\mu$ is the so-called inverse link function). Extending their algorithm to our setting, we give the full pseudocode of LogisticUCB in the Supplementary Materials section. We take all of the same preprocessing steps as in LinUCB, and we report the cumulative regret corresponding to the \textit{best} UCB constant. For the same reasons as above, we use regularized logistic regression in practice with cross-validation to tune the regularization parameter.
	
%
%
%
%
	
\item \textit{OfflineTree}. Recall that the simulated sports-advertising dataset was constructed using decision tree truths. Thus, it is meaningful to compare TreeBootstrap with a regret of zero, as it is possible in theory for estimated decision trees to capture the truth exactly.  However, as the four UCI datasets are all composed of real observations, there will always be part of the ``truth" which decision trees cannot capture. Our benchmark OfflineTree measures this error, serving as a meaningful \textit{lower bound} against which we can compare our algorithm. Described in detail in Algorithm \ref{alg:OfflineTree}, it can essentially be thought of as the offline classification error of decision trees with respect to a held-out test set. In our experimental results, we report the difference in cumulative regret on the UCI datasets between the candidate algorithms and OfflineTree.
	
			\begin{algorithm}
				\caption{OfflineTree()}
				For each observation $(x,y)$, define $w_a(x,y)$ as follows:\\ 
                \hspace{0.1in} $w_a(x,y) = 1$ if $y = a$, and \\
                \hspace{0.1in} $w_a(x,y) = 0$ if $y \neq a$ \\
				\vspace{0.05in}
				Hold out $T$ random observations, $\{x_t\}_{1 \leq t \leq T} $ \\
				\vspace{0.05in}
				
				\For{$a=1,...,K$}{
					Fit tree $\hat{\theta}_{a}$ on remaining data using $w_a(x,y)$ as the response variable
				}
				
				\For{$t=1,...,T$}{
					\vspace{0.2mm}
					Observe context vector $x_t$\\
					\vspace{0.8mm}
					
					Choose action $a_t = \displaystyle{\argmax_{a} \  \hat{p}(\hat{\theta}_{a},x_t)}$\\
				}  \label{alg:OfflineTree}
				
			\end{algorithm}
	
\end{enumerate}

\subsection{SPORTS ADVERTISING DATASET}

First, TreeBootstrap was tested under an idealistic setting -- a simulated dataset where the context-reward model is a decision tree for each action. We frame this dataset in the context of sports web-advertising. Whenever a user visits the website, we must offer ads for one of $K=4$ different products: golf clubs, basketball nets, tennis rackets, and soccer balls. We receive a reward of 1 if the user clicks the ad; otherwise, we receive no reward. 

\begin{figure}[h]
	\center{\includegraphics[trim={0 0 0 1.5cm},clip,width=0.45\textwidth]{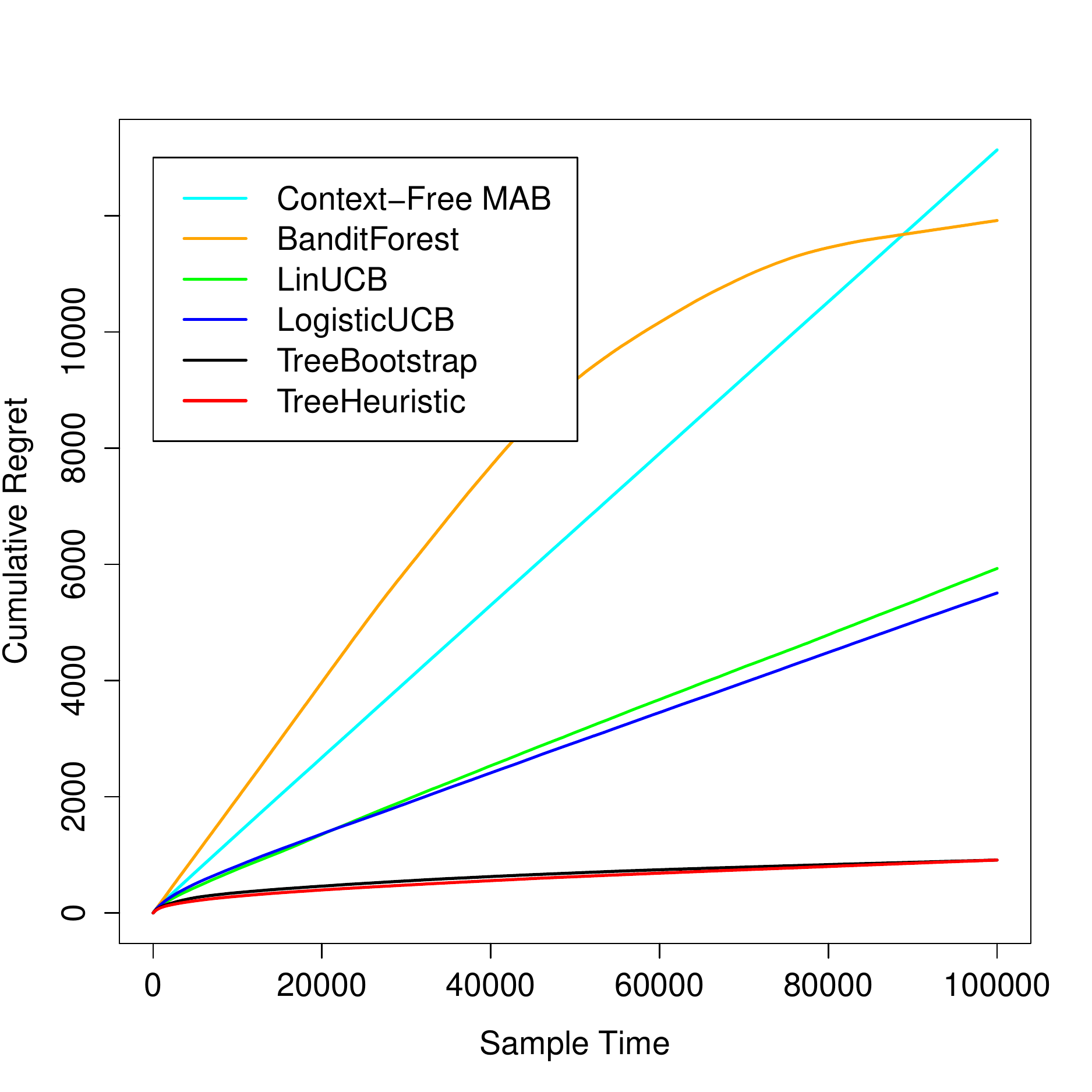}}
	\caption{Cumulative regret incurred on the (simulated) sports advertising dataset. 
    }
	\label{fig:SimulatedResults}
\end{figure}

Figure \ref{fig:DecisionTree} provides an example of the decision tree used to simulate the golf advertisement rewards, as well as information about the $M=4$ (binary) contextual variables available for each user. Figure \ref{fig:SimulatedResults} provides the cumulative regret data for the tested methods. As expected, our algorithm outperforms the other benchmark methods which do not use decision tree learners, and the performance of TreeBootstrap and TreeHeuristic are very similar. Moreover, the regret seems to converge to zero for our decision tree algorithms as the number of observed users becomes large. Finally, note how BanditForest eventually achieves a regret comparable to the other algorithms, but nonetheless incurs a much higher cumulative regret. This is due to the fact that the algorithm takes most of the time horizon to exit its ``pure exploration" phase, an observation which also holds across all the UCI datasets.

\subsection{UCI REPOSITORY DATASETS}

We next evaluated our algorithm on four classification datasets from the UCI Machine Learning Repository \citep{Lichman:2013}: \textit{Adult}, \textit{Statlog (Shuttle)}, \textit{Covertype}, and \textit{US Census Data (1990)}. The response variables used were \textit{occupation}, \textit{Cover\_Type}, and \textit{dOccup} for \textit{Adult}, \textit{Covertype}, and \textit{Census}, respectively, while for \textit{Shuttle} we used the variable corresponding to the last column of the dataset. We first ran a preprocessing step removing classes which were significantly underrepresented in each dataset (i.e., less than $0.05\%$ of the observations). After preprocessing, these datasets had $K = 12, 4, 7,$ and $9$ classes, respectively, as well as $M = 14, 9, 54,$ and $67$ contextual variables. We then constructed a bandit problem from the data in the following way: a regret of 0 (otherwise 1) is incurred if and only if the algorithm predicts the class of the data point correctly. This framework for adapting classification data for use in bandit problems is commonly used in the literature \citep{allesiardo2014neural,hsu2014taming,feraud2016random}. 

\begin{figure*}
	\begin{center}
	\subfloat[\textit{Adult} Dataset Results]{\includegraphics[width=0.45\textwidth]{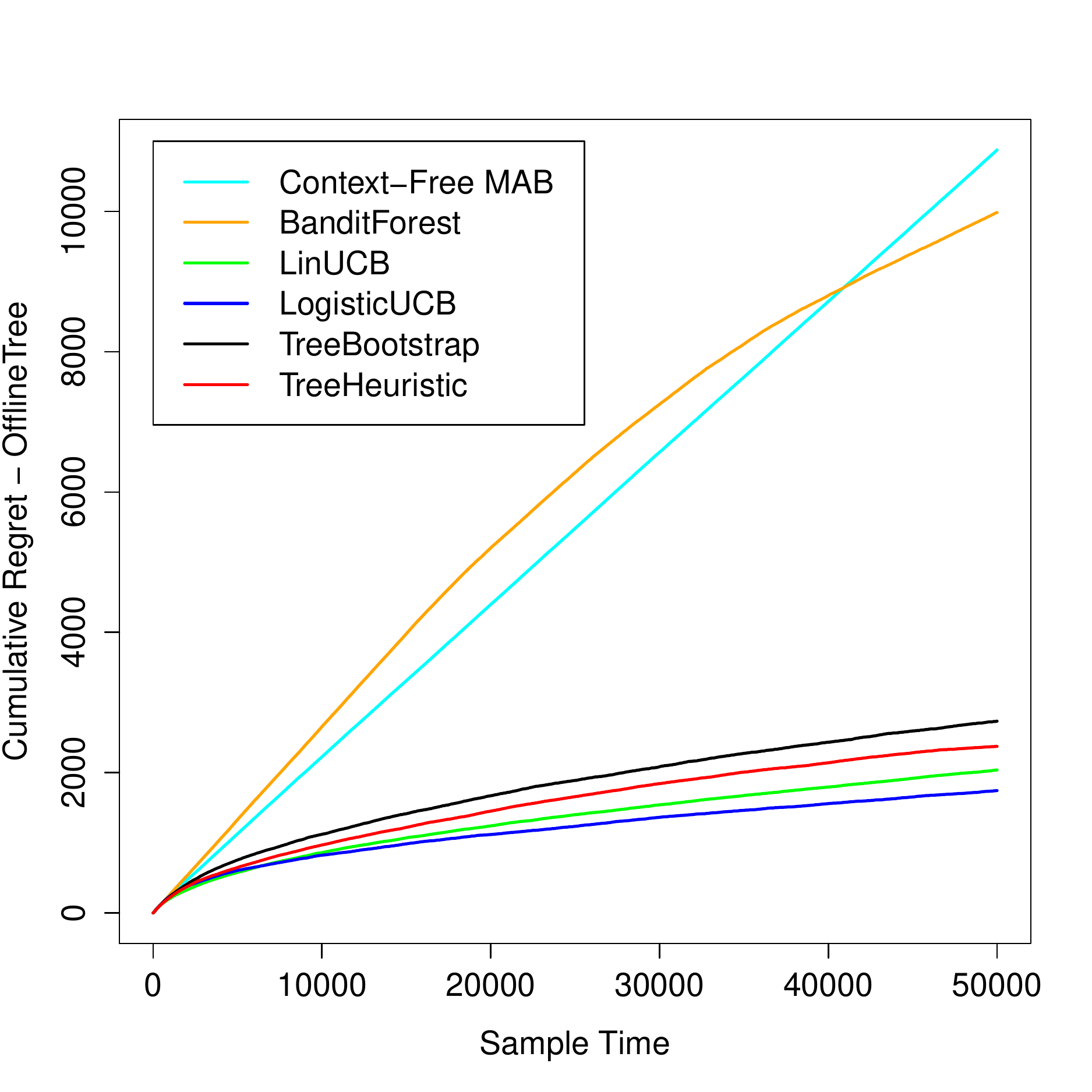}} 
	\subfloat[\textit{Statlog (Shuttle)} Dataset Results]{\includegraphics[width=0.45\textwidth]{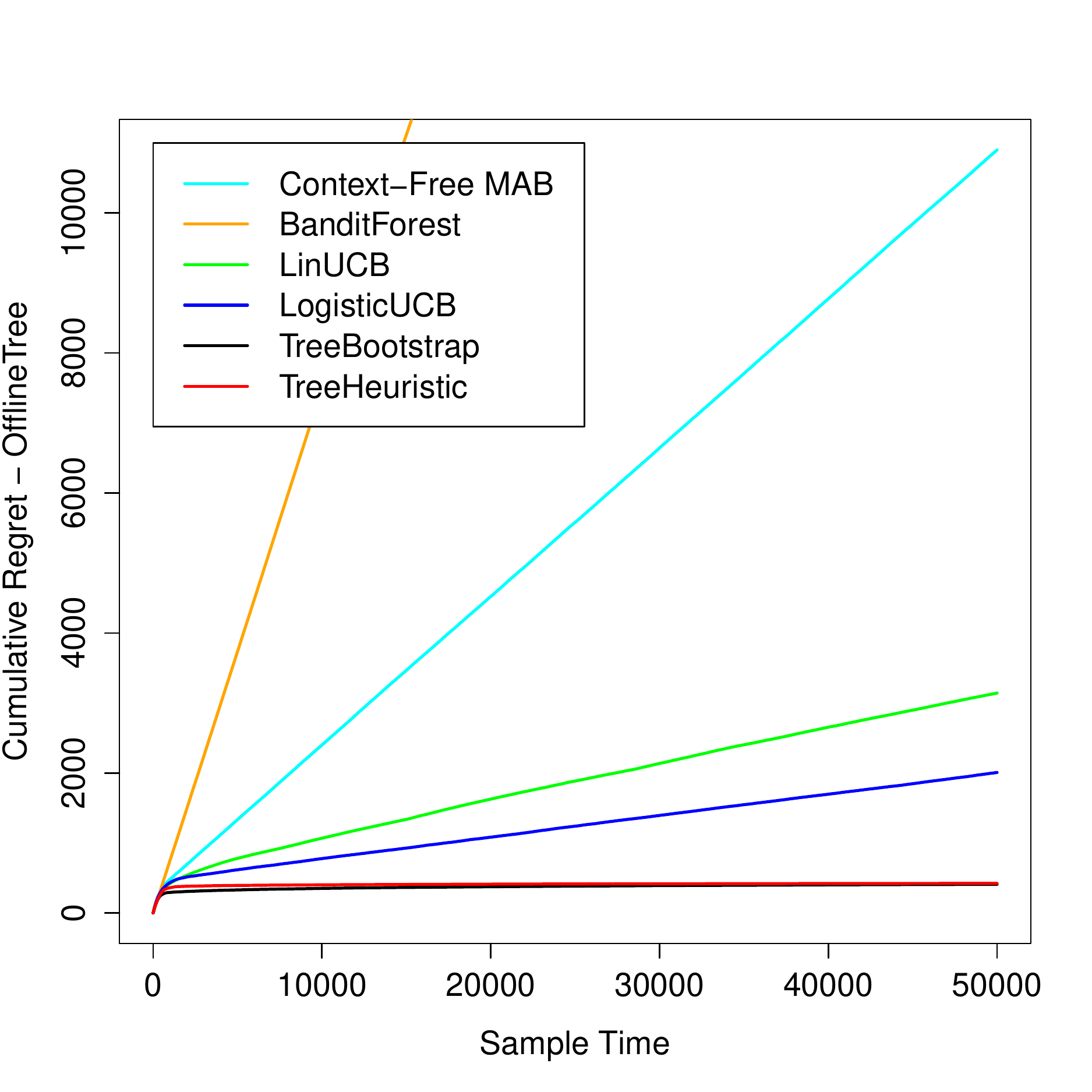}}\ \subfloat[\textit{Covertype} Dataset Results]{\includegraphics[width=0.45\textwidth]{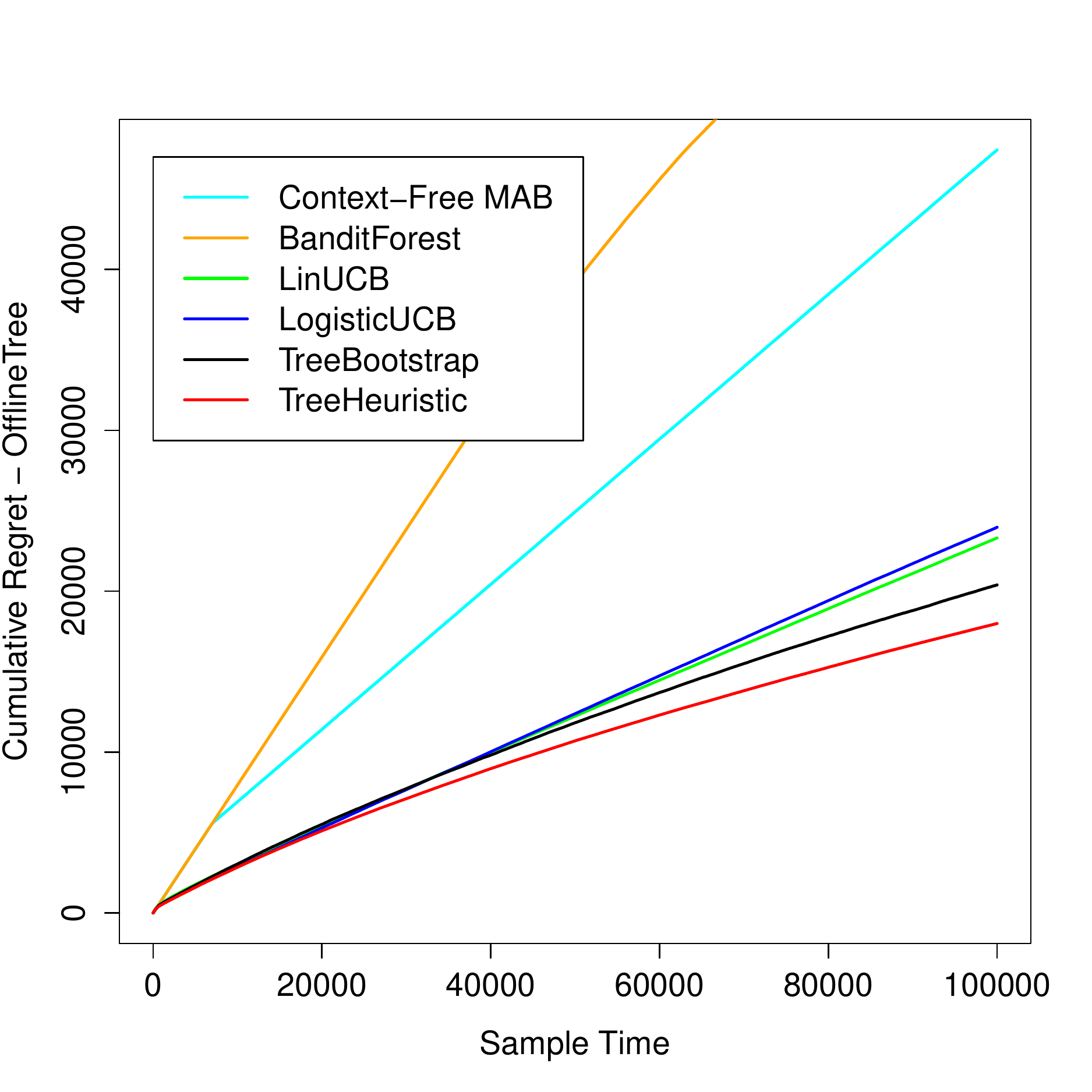}}
	\subfloat[\textit{Census} Dataset Results]{\includegraphics[width=0.45\textwidth]{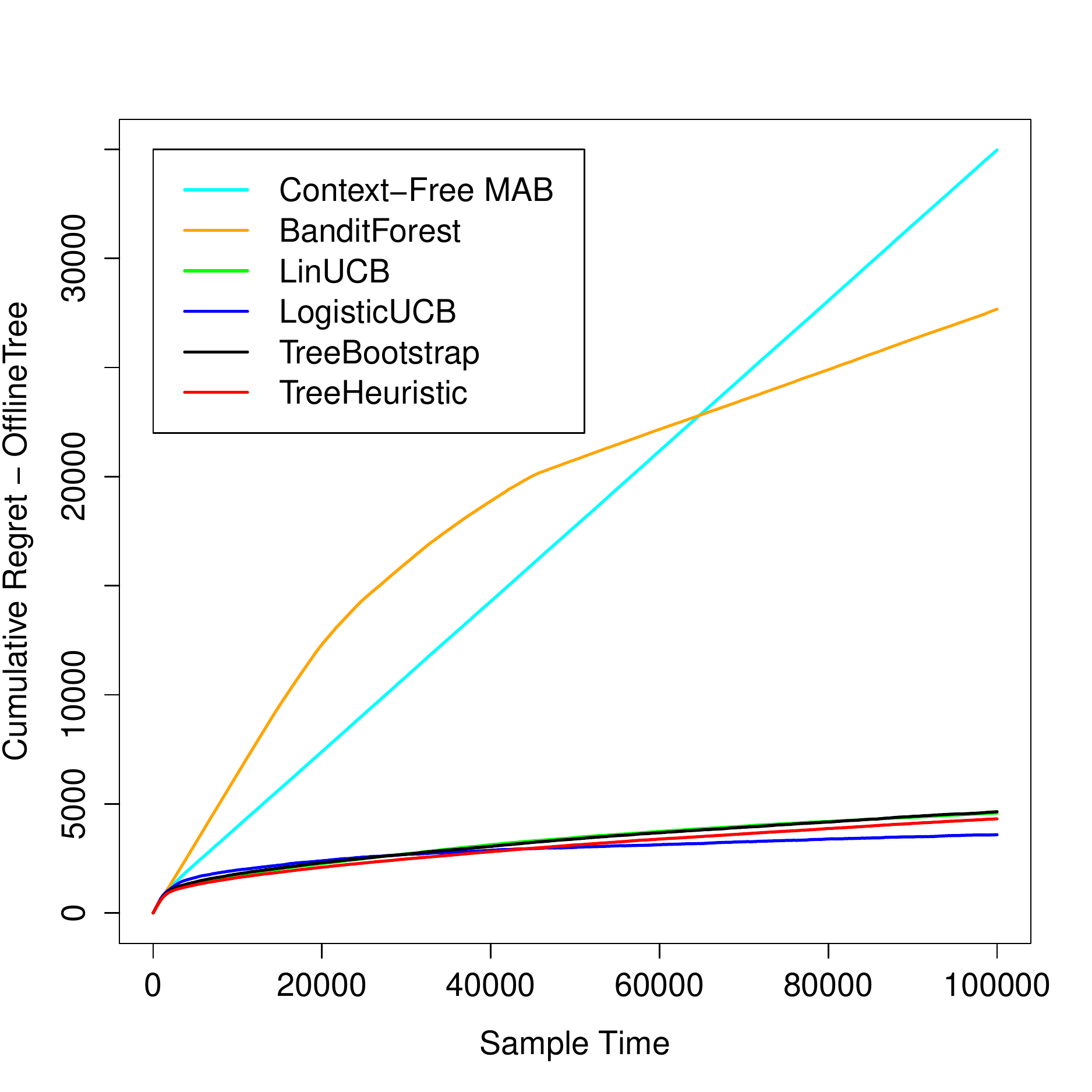}} 
	\quad \\
	\end{center}
	\caption{Cumulative regret incurred on various classification datasets from the UCI Machine Learning Repository. A regret of 0 (otherwise 1) is incurred iff a candidate algorithm predicts the class of the data point correctly.} \label{fig:UCIResults}
	
\end{figure*}

Figure \ref{fig:UCIResults} shows the cumulative regret of TreeBootstrap compared with the benchmarks on the UCI datasets. Recall that we plot regret relative to OfflineTree. In all cases, our heuristic achieves a performance equal to or better than that of TreeBootstrap. Note that LogisticUCB outperforms LinUCB on all datasets except \textit{Covertype}, which demonstrates the value of using learners in our setting which handle binary response data effectively.

\begin{figure}[h]
	\center{\includegraphics[width=0.45\textwidth]{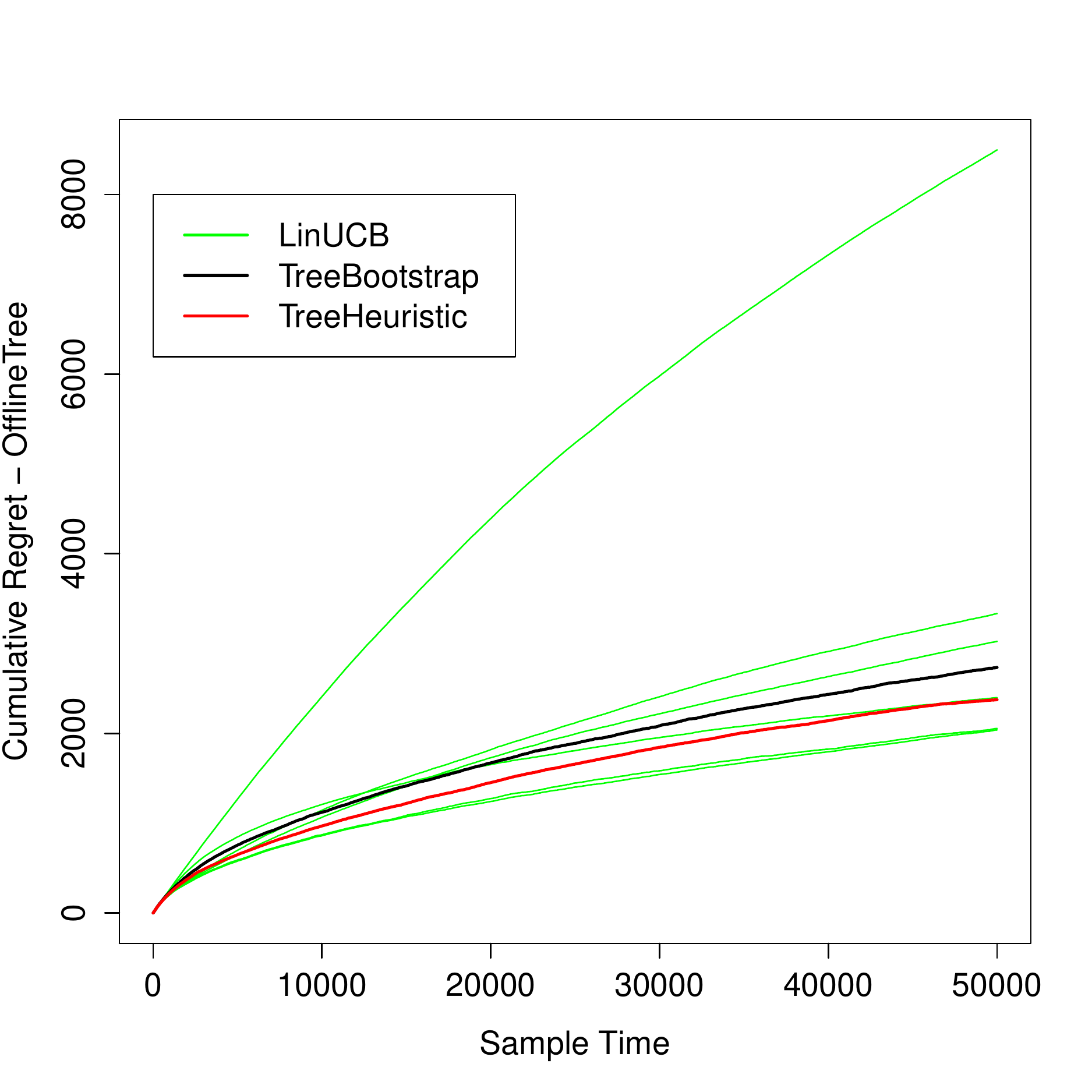}}
	\caption{Cumulative regret incurred on the \textit{Adult} dataset. The performance of LinUCB is given with respect to all values of the tuneable parameter attempted: $\alpha$ = 0.0001, 0.001, 0.01, 0.1, 1, 10}
	\label{fig:AdultLinUCBTuning}
\end{figure}

LinUCB and LogisticUCB outperform our algorithm on two of the four UCI datasets (\textit{Adult} and \textit{Census}). However, there are several caveats to this result. First, recall that we only report the cumulative regret of LinUCB and LogisticUCB with respect to the best exploration parameter, which is impossible to know a priori. Figure \ref{fig:AdultLinUCBTuning} shows LinUCB's cumulative regret curves corresponding to each value of the exploration parameter $\alpha$ implemented on the \textit{Adult} dataset. We overlay this on a plot of the cumulative regret curves associated with TreeBootstrap and our heuristic. Note that TreeBootstrap and TreeHeuristic outperformed LinUCB in at least half of the parameter settings attempted. Second, the difference in cumulative regret between TreeBootstrap and Linear/Logistic UCB on \textit{Adult} and \textit{Census} appears to approach a constant as the time horizon increases. This is due to the fact that decision trees will capture the truth eventually given enough training data. Conversely, in settings such as the sports ad dataset, \textit{Covertype}, and \textit{Shuttle}, it appears that the linear/logistic regression models have already converged and fail to capture the context-reward distribution accurately. This is most likely due to the fact that feature engineering is needed for the data to satisfy the GLM framework. Thus, the difference in cumulative regret between Linear/Logistic UCB and TreeBootstrap will become arbitrarily large as the time horizon increases. Finally, note that we introduce regularization into the linear and logistic regressions, tuned using cross-validation, which improve upon the original framework for LinUCB and Logistic UCB.

\section{CONCLUSION}
We propose a contextual bandit algorithm, TreeBootstrap, which can be easily and effectively applied in practice. We use decision trees as our base learner, and we handle the exploration-exploitation trade-off using bootstrapping in a way which approximates the behavior of Thompson sampling. As our algorithm requires fitting multiple trees at each time step, we provide a computational heuristic which works well in practice. Empirically, our methods' performance is quite competitive and robust compared to several well-known algorithms.

\clearpage

\bibliography{mybib.bib}{}
\bibliographystyle{apalike}

\end{document}


{ \center \LARGE{
			Supplementary Materials For ``A Practical Method for Solving Contextual Bandit Problems Using Decision Trees''
		}}
		
		\section{Proof of Theorem 1}
		
For the standard multi-armed bandit problem, let $n_k$ be the number of total observed responses for action $k$ and $p_k$ be the observed success rate for action $k$ (proportion of success responses out of total observed responses). For simplicity, we assume that we have observed at least one success and one failure for each action, i.e. $n_k \geq 2$ and $p_k \in (0,1)$ for all $k$. Then, the following theorem holds:
\begin{theorem}
	Let $a_t^{TS}$ be the action chosen by the Thompson sampling algorithm, and let $a_t^{B}$ be the action chosen by the bootstrapping algorithm given data $(n_k,p_k)$ for each action $k \in \{1,2,\ldots,K\}$. Then, 
	\[
	| P(a_t^{TS} = k) - P(a_t^{B} = k) | \leq C_k(p_1,...,p_K) \sum_{j=1}^K \frac{1}{\sqrt{n_j}}
	\]
	holds for every $k \in \{1,2,\ldots,K\}$, for some function $C_k(p_1,...,p_K)$ of $p_1,...,p_K$. 
\end{theorem}

To prove the theorem, we first provide a few lemmas. The proofs for all lemmas are given in Section \ref{sec:Lemmas} of this supplementary materials document. For notational convenience, we define $\alpha_k = n_k p_k$ and $\beta_k = n_k (1-p_k)$, which indicate the number of success and failure responses observed so far for action $k$. For each action $k \in \{1,2,\ldots,K\}$, the Thompson sampling algorithm first draws a random sample of the true (unknown) success probability according to a beta distribution with parameters $\alpha_k$ and $\beta_k$, and it then chooses the action with the highest success probability. The bootstrapping algorithm samples $n_k$ observations with replacement from action $k$'s observed rewards, and the generated success probability is then the proportion of successes observed in the bootstrapped dataset. This procedure is equivalent to generating a binomial random variable with $n_k$ trials and $p_k$ success rate, divided by $n_k$. The following lemma shows how close the distributions of these two random probability estimates are in terms of the number of available data points for each action. 
\begin{lemma}
	Let $X$ be a beta random variable with integer parameters $\alpha > 0$ and $\beta > 0$, and let $Y$ be a binomial random variable with $n$ trials and success rate $p$, where $\alpha + \beta = n$ and $p = \frac{\alpha}{n}$. Then,
	\[
	\max_{z \in [0,1]} \bigg| P(X \leq z) - P\left(\frac{Y}{n} \leq z\right) \bigg| \leq \frac{c(p)}{\sqrt{n}}\,,
	\]
	for some function $c(p)$ that is independent of $n$. 
	\label{lem:BetaBinomBound}
\end{lemma}

We now provide a second lemma which will prove useful in deriving Theorem 1. Recall that for each $k \in \{1,2,\ldots,K\}$, $n_k$ is the total number of observations and $p_k$ is the observed success rate. Let $p_k^{TS}$ be the randomly drawn success probability of action $k$ by the Thompson sampling algorithm and let $p_k^{B}$ be the randomly drawn success probability of action $k$ by the bootstrapping algorithm. Under each algorithm $h \in \{TS, B\}$, an action is chosen arbitrarily from the set $M^h := \{k: p_k^{h} \geq p_j^{h} \text{ for every } j\neq k\}$. In the Thompson sampling algorithm, $p_k^{TS}$ is sampled from a continuous (beta) distribution, and thus $M^{TS}$ will have cardinality one almost surely. Hence, 
\begin{eqnarray}
P(a_t^{TS} = k) = P(p_k^{TS} \geq p_j^{TS} \text{ for every } j\neq k)\,. \label{eqn:thompson}
\end{eqnarray}
However, these events are not necessarily equivalent with respect to the bootstrapping algorithm. Since $p_k^{B}$ is sampled using a discrete (binomial) distribution, it is possible that two actions will have the same sampled probabilities. Thus, it is possible for an action $k \in M^{B} $ to not be chosen if there exists another action $l \in M^{B}$. We provide a lemma which examines the difference in these two events: 

\begin{lemma}
	For some function $b(p_k)$ that is independent of $n_k$,
	\[
	|P(a_t^B = k) - P(p_k^{B} \geq p_j^{B} \textup{ for every } j\neq k)| \leq  \frac{b(p_k)}{\sqrt{n_k}} (1+O(1/n_k) )\,.
	\]
	\label{lem:BinomRepeatsBound}
\end{lemma} 

\vspace*{-1cm}

We can now proceed with the derivation of the theorem. We define $err_k(z|X) := P(p_k^{TS} \leq z | X) - P(p_k^{B} \leq z | X)$ with respect to some random variable $X$. Assuming $X$ is independent of $p_k^{TS}$ and $p_k^{B}$ (but not necessarily of $z$), then it follows from Lemma \ref{lem:BetaBinomBound} that $|err_k(z|X) |\leq \frac{c(p_k)}{\sqrt{n_k}}$ for some function $c(p_k)$ that is independent of $n_k$. Note that this result also holds for the function $\widetilde{err}_k(z|X) := P(p_k^{B} \leq z | X) - P(p_k^{TS} \leq z | X)$. 

Note that the events $\{ p_k^h \geq p_j^h \}$ for $j \neq k$ are independent conditioned on $p_k^h$. Hence, using (\ref{eqn:thompson}) we have 
\begin{eqnarray}
P(a_t^{TS} = k) & =&  E_{p_k^{TS}}\left[ P(p_k^{TS} \geq p_j^{TS} \text{ for every } j\neq k | p_k^{TS}) \right] \nonumber \\
& = & E_{p_k^{TS}}\left[ \prod_{j \neq k} P(p_k^{TS} \geq p_j^{TS} | p_k^{TS}) \right]  \nonumber \\
& = & E_{p_k^{TS}}\left[ \prod_{j \neq k} \left( P(p_k^{TS} \geq p_j^{B} | p_k^{TS}) + err_j(p_k^{TS} | p_k^{TS}) \right) \right]\,. \label{eqn:product}
\end{eqnarray}
In the expansion of the product (\ref{eqn:product}), only one term does not include $err_j(p_k^{TS}| p_k^{TS})$. This term is given as 
\begin{eqnarray}
&& E_{p_k^{TS}}\left[ \prod_{j \neq k} P(p_k^{TS} \geq p_j^{B} | p_k^{TS}) \right] \nonumber \\ 
& = & E_{p_k^{TS}}\left[ P(p_k^{TS} \geq p_j^{B} \text{ for every } j\neq k | p_k^{TS}) \right] \nonumber \\
& = & E_{p_j^{B},j \neq k}\left[ P(p_k^{TS} \geq \max_{j \neq k} p_j^{B} | p_j^{B},j \neq k) \right] \nonumber \\
& = & E_{p_j^{B},j \neq k}\left[ P(p_k^{B} \geq \max_{j \neq k} p_j^{B} | p_j^{B},j \neq k) + \widetilde{err}_k(\max_{j \neq k} p_j^{B}|p_j^{B},j \neq k) \right] \nonumber \\
& = & P(p_k^{B} \geq \max_{j \neq k} p_j^{B}) + E_{p_j^{B},j \neq k}[\widetilde{err}_k(\max_{j \neq k} p_j^{B}|p_j^{B},j \neq k)] \nonumber \\
& = & P(a_t^B = k) + (P(p_k^{B} \geq \max_{j \neq k} p_j^{B}) - P(a_t^B = k))+ E_{p_j^{B},j \neq k}[\widetilde{err}_k(\max_{j \neq k} p_j^{B}|p_j^{B},j \neq k)]\,. \label{eqn:applied}
\end{eqnarray}
The rest of (\ref{eqn:product}) is the sum of multiplications of $K-1$ terms of $P(p_k^{TS} \geq p_j^{B}|p_k^{TS})$ and $err_j(p_k^{TS}|p_k^{TS})$. Because $P(p_k^{TS} \geq p_j^{B}|p_k^{TS}) \leq 1$ and $|err_j| \leq \frac{c(p_j)}{\sqrt{n_j}}$ from Lemma \ref{lem:BetaBinomBound}, the rest can be bounded by $\prod_{j \neq k} \left (1+ \frac{c(p_j)}{\sqrt{n_j}} \right ) -1$. Hence, applying Lemma \ref{lem:BetaBinomBound} and Lemma \ref{lem:BinomRepeatsBound} to (\ref{eqn:applied}), we have 
\begin{eqnarray}
&&| P(a_t^{TS} = k) - P(a_t^{B} = k) | \nonumber \\ & \leq &  |P(p_k^{B} \geq \max_{j \neq k} p_j^{B}) - P(a_t^B = k)|+ |E_{p_j^{B},j \neq k}[\widetilde{err}_k(\max_{j \neq k} p_j^{B}|p_j^{B},j \neq k)]|+ \prod_{j \neq k} \left (1+ \frac{c(p_j)}{\sqrt{n_j}} \right ) -1 \nonumber \\
& \leq & \frac{b(p_k)}{\sqrt{n_k}} \left (1+O(1/n_k) \right )+ \frac{c(p_k)}{\sqrt{n_k}} + \prod_{j \neq k} \left (1+ \frac{c(p_j)}{\sqrt{n_j}} \right ) -1\,.
\label{eqn:almostthere}
\end{eqnarray}

Note that the error term from Sterling's approximation, $O(1/n_k)$, can be bounded from above by a constant. Furthermore, for any set $S$ of actions, 
\begin{eqnarray*}
	\prod_{s \in S} \frac{c(p_s)}{\sqrt{n_s}} \leq \frac{1}{\sqrt{n_t}}\prod_{s \in S} c(p_s) \qquad \forall t \in S\,.
\end{eqnarray*}

One can then use these facts to manipulate (\ref{eqn:almostthere}) to prove the desired result:
\begin{eqnarray*}
	| P(a_t^{TS} = k) - P(a_t^{B} = k) | \leq C_k(p_1,...,p_K) \sum_{j=1}^K \frac{1}{\sqrt{n_j}}\,,
\end{eqnarray*}

where $C_k(p_1,...,p_K)$ is a function of $p_1,...,p_K$.  \qed

\newpage

\section{Proof of Auxiliary Lemmas} \label{sec:Lemmas}

\setcounter{lemma}{0}

\begin{lemma}
	Let $X$ be a beta random variable with integer parameters $\alpha > 0$ and $\beta > 0$, and let $Y$ be a binomial random variable with $n$ trials and success rate $p$, where $\alpha + \beta = n$ and $p = \frac{\alpha}{n}$. Then,
	\[
	\max_{z \in [0,1]} \bigg| P(X \leq z) - P\left(\frac{Y}{n} \leq z\right) \bigg| \leq \frac{c(p)}{\sqrt{n}}\,,
	\]
	for some function $c(p)$ that is independent of $n$. 
	\label{lem:BetaBinomBound}
\end{lemma}
\proof{}
For notational convenience, we define $q = \frac{\beta}{n} = 1-p$, and we denote the p.d.f., and the c.d.f. of a standard normal random variable by $\phi(\cdot)$ and $\Phi(\cdot)$, respectively. We will first show that, for each $z \in [0,1]$, $P(X \leq z)$ and  $P\left(\frac{Y}{n} \leq z\right)$ can be approximated using normal c.d.f.'s  $\Phi( \frac{\sqrt{n}(z-p)}{\sqrt{pq + (z-p)^2}})$ and  $\Phi(\frac{\sqrt{n}(z-p)}{\sqrt{pq}})$, respectively. Finally, we will bound the difference in these approximations and apply the triangle inequality:

\begin{eqnarray}
\bigg| P(X \leq z) - P\left(\frac{Y}{n} \leq z\right) \bigg| & \leq & \bigg| P(X \leq z) - \Phi( \frac{\sqrt{n}(z-p)}{\sqrt{pq + (z-p)^2}}) \bigg| \nonumber \\
& + &  \bigg| \Phi(\frac{\sqrt{n}(z-p)}{\sqrt{pq}})- \Phi(\frac{\sqrt{n}(z-p)}{\sqrt{pq + (z-p)^2}}) \bigg| \nonumber \\ 
& + & \bigg| P(\frac{Y}{n} \leq z) - \Phi(\frac{\sqrt{n}(z-p)}{\sqrt{pq}}) \bigg|  \label{eqn:triangle}
\end{eqnarray}

We first bound the normal approximation for  $P\left(\frac{Y}{n} \leq z\right)$. From the Berry - Esseen theorem, we have 
\[ | P( \sqrt{n} \frac{(Y/n - p)}{\sqrt{pq}} \leq x ) - \Phi(x) | \leq \frac{C(p^2 + q^2)}{\sqrt{npq}}\] for every $x$, 
which implies that 
\begin{equation}
| P(\frac{Y}{n} \leq z) - \Phi(\frac{\sqrt{n}(z-p)}{\sqrt{pq}})| \leq \frac{C(p^2 + q^2)}{\sqrt{npq}} \label{eqn:bound_bino}
\end{equation}
holds for every $z \in [0,1]$. 

Next, we will show that $P(X \leq z)$ can be approximated by a similar (but not exactly the same) function, $\Phi( \frac{\sqrt{n}(z-p)}{\sqrt{pq + (z-p)^2}})$. Note that the beta random variable $X$ has the same distribution as $\frac{A}{A + B}$, where $A$ and $B$ are independent Gamma random variables with shape parameters $\alpha$ and $\beta$, respectively. We first derive an approximation for the c.d.f. of Gamma random variables. Suppose that $\Gamma$ is Gamma distributed with an integer shape parameter $m > 0$. Because $\Gamma$ has the same distribution as the sum of $m$ independent exponential random variables with parameter $1$, from the Berry - Esseen theorem we have
\begin{equation}  |P( \sqrt{m} (\Gamma/m - 1) \leq x) - \Phi(x) | \leq  \frac{C\rho}{\sqrt{m}}\,, \label{eqn:gamma}
\end{equation} 
where $\rho$ is the third-order absolute moment of the unit exponential distribution. 

Let $N_1$ and $N_2$ be independent standard normal random variables, and define 
\begin{eqnarray*}
	g(z) := P( (\sqrt{\alpha}N_1+\alpha)(1-z) \leq  (\sqrt{\beta} N_2+\beta) z )\,.
\end{eqnarray*}
Then, from the triangle inequality we have that for all $z \in (0,1)$:
\begin{eqnarray*}
	& & | P(X \leq z) - g(z) | = | P(A(1-z) \leq Bz) - g(z)| \\
	& \leq & | P(A(1-z) \leq Bz) -  P((\sqrt{\alpha}N_1+\alpha)(1-z) \leq Bz)| + | P((\sqrt{\alpha}N_1+\alpha)(1-z) \leq Bz) - g(z)|\,. 
\end{eqnarray*}
From (\ref{eqn:gamma}), we have
\begin{eqnarray*}
	& & | P(A(1-z) \leq Bz) -  P((\sqrt{\alpha}N_1+\alpha)(1-z) \leq Bz)| \\
	& = & \big| E_B\left[ P(A(1-z) \leq Bz | B) - P((\sqrt{\alpha}N_1+\alpha)(1-z) \leq Bz | B) \right] \big| \\
	& \leq & E_B\left[ \big| P(A(1-z) \leq Bz | B) - P((\sqrt{\alpha}N_1+\alpha)(1-z) \leq Bz | B) \big| \right]  \\
	& \leq & \frac{C\rho}{\sqrt{\alpha}}\,.
\end{eqnarray*}
Similarly, again from (\ref{eqn:gamma}), we have
\begin{eqnarray*}
	& & |P((\sqrt{\alpha}N_1+\alpha)(1-z) \leq Bz) - g(z)| \\
	& = & \big| E_{N_1}\left[P((\sqrt{\alpha}N_1+\alpha)(1-z) \leq Bz|N_1) - P( \sqrt{\alpha}N_1(1-z) -  \sqrt{\beta}N_2 z \leq nz - \alpha|N_1)\right]\big|\\
	& \leq &  E_{N_1}\left[ \big| P((\sqrt{\alpha}N_1+\alpha)(1-z) \leq Bz|N_1) - P( \sqrt{\alpha}N_1(1-z) -  \sqrt{\beta}N_2 z \leq nz - \alpha|N_1) \big| \right]\\
	& \leq & \frac{C\rho}{\sqrt{\beta}}\,.
\end{eqnarray*}
Finally, because $-N_2$ is a standard normal random variable and the sum of two independent normal random variables is a standard normal random variable, we have 
\begin{eqnarray*}
	& & g(z) = P( \sqrt{\alpha}N_1(1-z) -  \sqrt{\beta}N_2 z \leq nz - \alpha) \\
	& = & \Phi(\frac{nz - \alpha}{\sqrt{\alpha(1-z)^2 + \beta z^2}})  = \Phi( \frac{\sqrt{n}(z-p)}{\sqrt{pq + (z-p)^2}})\,, 
\end{eqnarray*}
which concludes that 
\begin{equation}
|P(X \leq z) -  \Phi( \frac{\sqrt{n}(z-p)}{\sqrt{pq + (z-p)^2}}) | \leq \frac{C\rho}{\sqrt{np}} +\frac{C\rho}{\sqrt{nq}} \label{eqn:bound_beta}
\end{equation}
holds for every $z \in [0,1]$. 

Finally, we provide a bound for $| \Phi(\frac{\sqrt{n}(z-p)}{\sqrt{pq}})- \Phi(\frac{\sqrt{n}(z-p)}{\sqrt{pq + (z-p)^2}})|$. For notational convenience, we define $d := \frac{z-p}{\sqrt{pq}}$. Then, the difference is given as $|\Phi(\sqrt{n}d) - \Phi(\frac{\sqrt{n}d}{\sqrt{1+d^2}})|$. Because $\Phi(y)$ is concave in $y$ for $y \geq 0$, for every $d \geq 0$ we have 
\begin{eqnarray*}
	|\Phi(\sqrt{n}d) - \Phi(\frac{\sqrt{n}d}{\sqrt{1+d^2}})|&=& \Phi(\sqrt{n}d) - \Phi(\frac{\sqrt{n}d}{\sqrt{1+d^2}})  \\
	& \leq &  \left(\sqrt{n}d - \frac{\sqrt{n}d}{\sqrt{1+d^2}}\right) \phi(\frac{\sqrt{n}d}{\sqrt{1+d^2}}) \\
	& = & (\sqrt{1+d^2}-1)\frac{\sqrt{n}d}{\sqrt{1+d^2}} \phi(\frac{\sqrt{n}d}{\sqrt{1+d^2}})  \\
	& \leq & (\sqrt{1+d^2} - 1)\phi(1) \\
	& = & \left(\frac{d^2}{\sqrt{1+d^2} + 1}\right)\phi(1) \\ &\leq& \frac{d^2}{2}\phi(1)\,,
\end{eqnarray*}
where the first inequality is from the concavity of $\Phi(\cdot)$ and the second inequality is from the fact that $x\phi(x)$ is maximized at $x = 1$ for $x \geq 0$. 
We define $f(d) := \Phi(\sqrt{n}d) - \Phi(\frac{\sqrt{n}d}{\sqrt{1+d^2}})$ and $d^*(n) := \argmax_{d \geq 0} f(d)$. Then, for every $d \geq 0$, $\Phi(\sqrt{n}d) - \Phi(\frac{\sqrt{n}d}{\sqrt{1+d^2}}) \leq \Phi(\sqrt{n}d^*(n)) - \Phi(\frac{\sqrt{n}d^*(n)}{\sqrt{1+d^*(n)^2}}) \leq  \frac{d^*(n)^2}{2}\phi(1)$. 

We will later show that $d^*(n)^4 < \frac{30 \ln(10)}{n}$. Hence, 
\begin{equation}
|\Phi(\frac{\sqrt{n}(z-p)}{\sqrt{pq}})- \Phi(\frac{\sqrt{n}(z-p)}{\sqrt{pq + (z-p)^2}}) |\leq \frac{\phi(1)\sqrt{30\ln(10)}}{2\sqrt{n}} \label{eqn:bound_cdf}
\end{equation}
holds for every $z \geq p$. The case of $z < p$ can be shown via symmetry. 

Finally, from applying the bounds (\ref{eqn:bound_bino}), (\ref{eqn:bound_beta}), and (\ref{eqn:bound_cdf}) to equation (\ref{eqn:triangle}), we have that
\[
\bigg| P(X \leq z) - P\left(\frac{Y}{n} \leq z\right) \bigg| \leq \frac{C(p^2 + q^2)}{\sqrt{npq}} +  \frac{C\rho}{\sqrt{np}} +  \frac{C\rho}{\sqrt{nq}} + \frac{\phi(1)\sqrt{30\ln(10)}}{2\sqrt{n}}\,,
\]
which proves the lemma. 

It remains to show that $d^*(n)^4 < \frac{30 \ln(10)}{n}$. The first-order condition for $d^*(n)$ is given as 
\begin{eqnarray*} 
	f'(d) = \sqrt{n}\phi(\sqrt{n}d) - \sqrt{n}(1+d^2)^{-\frac{3}{2}}\phi(\frac{\sqrt{n}d}{\sqrt{1+d^2}}) =0\,. \end{eqnarray*}
Note that $f(d)$ is nonnegative and differentiable on $d \geq 0$. Since $f(0) = 0$, then if $\lim_{d \rightarrow \infty} f'(d) \leq 0$ we can conclude that there exists a global maximizer of $f(d)$ over $d \geq 0$ which satisfies the first-order condition. $f'(d) \leq 0$ can be simplified as
\begin{eqnarray*}
	\exp\left( - \frac{n d^2}{2} + \frac{n d^2}{2(1+d^2)}\right) & \leq &(1+d^2)^{-\frac{3}{2}} \\
	-\frac{n d^4}{2(1+d^2)}& \leq & -\frac{3}{2} \ln(1+d^2) \\
	n& \geq &\frac{3(1+d^2)\ln(1+d^2)}{d^4}\,.
\end{eqnarray*}
Note that the right hand side approaches zero as $d \rightarrow \infty$, proving that $\lim_{d \rightarrow \infty} f'(d) \leq 0$. Thus, $d^*(n)$ satisfies the first-order condition, given in a simplified form below:
\begin{eqnarray*}
	n& = &\frac{3(1+d^2)\ln(1+d^2)}{d^4}\,.
\end{eqnarray*}
Note that 
\begin{eqnarray*}
	& & \frac{d}{d x}\left(\frac{(1+x)\ln(1+x)}{x^2} \right) = \frac{\ln(1+x)}{x^2} + \frac{1}{x^2} - 2 \frac{(1+x)\ln(1+x)}{x^3} \\
	& =  & \frac{1}{x^3}\left( x - (x+2)\ln(x+1)\right) \leq 0\,,
\end{eqnarray*}
because $(x - (x+2)\ln(x+1))$ is decreasing in $x$ and is zero when $x = 0$. Hence, $d^*(n)$ is decreasing in $n$. 

From the first-order condition and the decreasing property of $d^*(n)$, we have 
\[ d^*(n)^4 = \frac{3(1+d^*(n)^2)\ln(1+d^*(n)^2)}{n} \leq \frac{3(1+d^*(1)^2)\ln(1+d^*(1)^2)}{n}\,.
\] From the decreasing property and the first-order condition, we can show that $d^*(1) < 3$, which implies that $\frac{3(1+d^*(1)^2)\ln(1+d^*(1)^2)}{n} < \frac{30 \ln(10)}{n}$. \qed
\\

\begin{lemma}
	For some function $b(p_k)$ that is independent of $n_k$,
	\[
	|P(a_t^B = k) - P(p_k^{B} \geq p_j^{B} \textup{ for every } j\neq k)| \leq  \frac{b(p_k)}{\sqrt{n_k}} (1+O(1/n_k) )\,.
	\]
	\label{lem:BinomRepeatsBound}
\end{lemma}

\vspace*{-1cm}

\proof{} Note that
\begin{eqnarray*}
	P(p_k^{B} \geq p_j^{B} \text{ for every } j\neq k) & = & P(p_k^{B} \geq \max_{j \neq k} p_j^{B}) \\ 
	& = & P(p_k^{B} \geq \max_{j \neq k} p_j^{B}, a_t^B = k) + P(p_k^{B} \geq \max_{j \neq k} p_j^{B}, a_t^B \neq k) \\
	& = & P(a_t^B = k) + P(p_k^{B} \geq \max_{j \neq k} p_j^{B}, a_t^B \neq k)\,.
\end{eqnarray*}
Let $Y_k$ denote the binomial variable associated with $p_k^B$ for each action $k$, i.e. $Y_k := n_k p_k^B$. Then,
\begin{eqnarray*}
	|P(a_t^B = k) - P(p_k^{B} \geq \max_{j \neq k} p_j^{B})| & = & P(p_k^{B} \geq \max_{j \neq k} p_j^{B}, a_t^B \neq k) \\
	& \leq & P(p_k^{B} = \max_{j \neq k} p_j^{B}) \\
	& = & P(Y_k  = n_k \max_{j \neq k} p_j^{B}) \\
	& \leq & \max_i P(Y_k  = i) \\
	& = & \max_i \binom{n_k}{i} p_k^i (1-p_k)^{n_k - i}\,.
\end{eqnarray*}
One can show that the binomial p.d.f. with parameters $(n, p)$ is maximized when $i = \lfloor (n + 1)p \rfloor$. Using the fact that $n_k p_k$ is an integer and that $p_k \in (0,1)$, $\lfloor (n_k + 1)p_k \rfloor = \lfloor n_k p_k + p_k \rfloor = n_k p_k$. Further, through applying Stirling's formula, $n! = \sqrt{2 \pi n} \left (\frac{n}{e} \right )^n (1 + O(1/n))$, one obtains
\begin{eqnarray*}
	\binom{n_k}{n_k p_k} &=& \frac{n_k!}{(n_k p_k)! (n_k - n_k p_k)!}\\ 
	&=& \sqrt{\frac{n_k}{2\pi n_k p_k (n_k-n_k p_k)}}\left ( \frac{n_k}{n_k p_k}\right)^{n_k p_k} \left (\frac{n_k}{n_k-n_k p_k}\right )^{n_k-n_k p_k}(1 + O(1/n_k)) \\ 
	&=& \frac{1}{\sqrt{2\pi n_k p_k (1- p_k)}}\left ( \frac{1}{ p_k}\right)^{n_k p_k} \left (\frac{1}{1-p_k}\right )^{n_k-n_k p_k}(1 + O(1/n_k))\,.
\end{eqnarray*}
Thus, 
\begin{eqnarray*}
	&& \max_i \binom{n_k}{i} p_k^i (1-p_k)^{n_k - i} \\
	& = & \binom{n_k}{n_k p_k} p_k^{n_k p_k} (1-p_k)^{n_k - n_k p_k} \\
	& = & \frac{1}{\sqrt{2\pi n_k p_k (1- p_k)}}\left ( \frac{1}{ p_k}\right)^{n_k p_k} \left (\frac{1}{1-p_k}\right )^{n_k-n_k p_k}  p_k^{n_k p_k} (1-p_k)^{n_k - n_k p_k} (1 + O(1/n_k)) \\
	& = & \frac{1}{\sqrt{2 \pi n_k p_k (1 - p_k)}} (1 + O(1/n_k))\,,
\end{eqnarray*}
which proves the lemma. \qed

\newpage
		
		\section{Pseudocode for Logistic UCB}
		
		\setcounter{algocf}{-1}
		
		\begin{algorithm} 
			\caption{LogisticUCB()}
			Define $\mathcal{L}(y) = \frac{e^y}{1+e^{y}}$ \\
			Define $\rho(s) = \sqrt{M \log s \log (sT/\delta)}$ \\
			\quad \\
			
			Initialize $t_a = 0 , X_{t,a} = I_M \ \forall a = 1,...,K$ \\
			\quad \\
			
			\For{$t=1,...,T$}{
				Observe context vector $x_t$\\
				\For{$a=1,...,K$}{
					$UCB_{t,a} = \mathcal{L}(x_t^T\hat{\theta}_{t,a_t})+\rho(t_a) ||x_t||_{X_{t,a}^{-1}}$
				}
				Choose action $a_t = \displaystyle{\argmax_{a} \  UCB_{t,a}}$\\
				\quad \\
				
				Update $t_{a_t} = t_{a_t} + 1, X_{t+1,a_t} = X_{t,a_t} + x_t x_t^T$ \\
				Update $\hat{\theta}_{t,a_t}$ with $(x_t, r_{t,a_t})$ \\
			} \label{alg:LogisticUCB}
			
		\end{algorithm}